\def\year{2021}\relax
\documentclass[letterpaper]{article} 
\usepackage{aaai21}  
\usepackage{times}  
\usepackage{helvet} 
\usepackage{courier}  
\usepackage[hyphens]{url}  
\usepackage{graphicx} 
\usepackage{natbib}  
\usepackage{caption} 
\usepackage{amsmath}
\usepackage{amssymb}
\usepackage{amsfonts}
\usepackage[switch]{lineno}  
\usepackage{color}
\title{In-game Residential Home Planning via Visual Context-aware Global Relation Learning}
\author{
    Lijuan Liu\textsuperscript{\rm 1},
    Yin Yang\textsuperscript{\rm 2}, 
    Yi Yuan\textsuperscript{\rm 1}\thanks{Corresponding author},
    Tianjia Shao\textsuperscript{\rm 3}, 
    He Wang\textsuperscript{\rm 4},  
    Kun Zhou\textsuperscript{\rm 3}\\
}

\affiliations{ 
    \textsuperscript{\rm 1}NetEase Fuxi AI Lab \\
    \textsuperscript{\rm 2}School of Computing, Clemson University \\
    \textsuperscript{\rm 3}State Key Lab of CAD\&CG, Zhejiang University \\
    \textsuperscript{\rm 4}Leeds University \\
    {liulijuan, yuanyi}@corp.netease.com, yin5@clemson.edu, tjshao@zju.edu.cn, H.E.Wang@leeds.ac.uk, kunzhou@acm.org \\
}

\begin{document}
\maketitle

\begin{abstract}
In this paper, we propose an effective global relation learning algorithm to recommend an appropriate location of a building unit for in-game customization of residential home complex. Given a construction layout, we propose a visual context-aware graph generation network that learns the implicit global relations among the scene components and infers the location of a new building unit. The proposed network takes as input the scene graph and the corresponding top-view depth image. It provides the location recommendations for a newly-added building units by learning an auto-regressive edge distribution conditioned on existing scenes. We also introduce a global graph-image matching loss to enhance the awareness of essential geometry semantics of the site. Qualitative and quantitative experiments demonstrate that the recommended location well reflects the implicit spatial rules of components in the residential estates, and it is instructive and practical to locate the building units in the 3D scene of the complex construction.

\end{abstract}



\section{Introduction}
Customized residential complex design becomes a popular element in modern MMORPG games. This module allows players to virtually create personalized housing experiences with a comprehensive construction and design interface. For instance, a player could have a palace-like mansion with a carefully-shaped garden of pools and a greenhouse. Possessing such luxury housing is unlikely to be possible for most of us. Yet, it could enhance the feeling of belongingness and escalate the joyfulness during the gaming. On the downside, designing a residential housing complex is not ``as easy as pie'', which requires professional expertise and extensive experiences. Our answer to this dilemma is to resort to machine learning to prompt smart suggestions during the user interaction, which is similar to smart typing system that predicts the following word we will input.


\begin{figure}[t!]
  \centering
  \includegraphics[width=\linewidth]{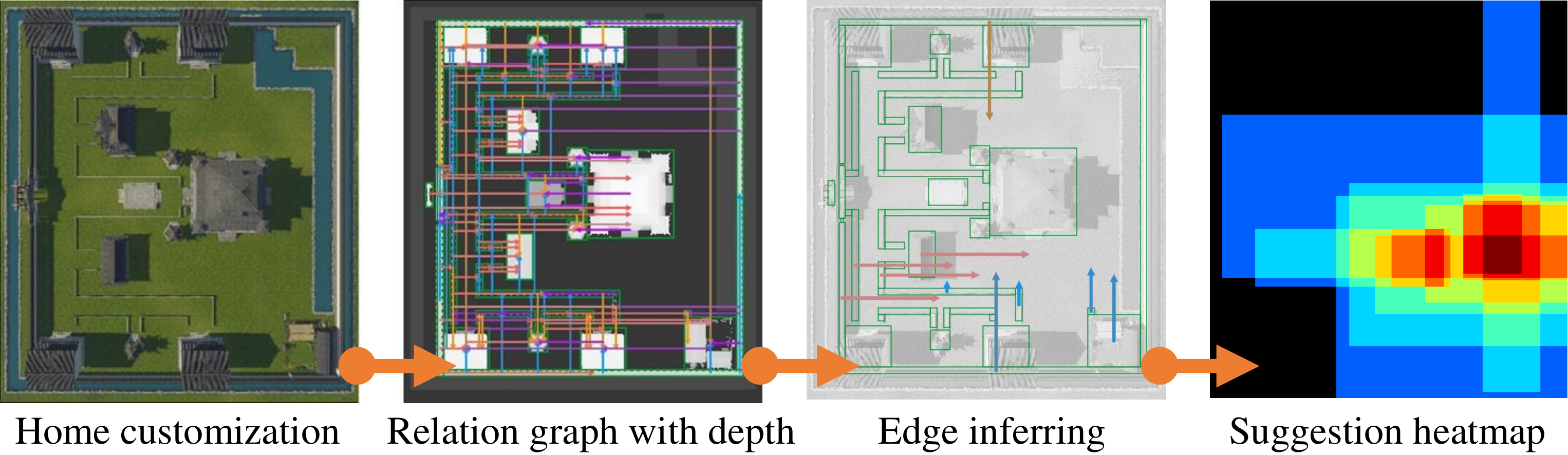}
  \caption{A high-level overview of our pipeline: from an input image of in-game the home customization site, our system extracts a relation graph. With other learned features, we train a graph generation network to infer the deployment of new edges. Finally, our system outputs a location prediction indicates the ``suitableness'' for the next building unit.}
  \label{fig:small-teaser}
\end{figure}

Following this motivation, we introduce an algorithm for location recommendation, which interactively provides suggestions to players on where to place new building components etc. A few techniques have been proposed to suggest a placement of a new component in an indoor scene automatically~\cite{wang2018deep, wang2019planit, nauata2020house, wu2019data}. They use deep learning techniques e.g., the FiLM net~\cite{perez2018film}, to predict a component location as an attribute of the new node. However, these approaches do not directly transfer to the home construction which occurs in an open outdoor space. Clearly, planning the building construction of a residential complex involves elements of diverse dimensions and scales. Therefore, searching all the possible positions on the site is not efficient. Second, items/elements in indoor scenes are normally associated with well-defined functional constraints, which can be fully exploited by the network. However, we have much weaker functional relations among buildings on the construction site. Certain types of building units are also exclusive -- for instance, one cannot add extra building blocks on the top of a swimming pool, and we name such locations \emph{forbidden areas}. This type of exclusiveness is not considered in previous algorithms.

Given a layout of a housing site, we aim to suggest the user a location directly without any prior knowledge of the building unit to be placed. Our method is inspired by planIT~\cite{wang2019planit} converting the site layout into a graph. To reduce the search space, we do not iterate all the candidate locations on the layout. Instead, we infer a possible location of a building unit through \emph{global} graph relations. While building units do not have strong local/neighborhood dependence, in a complex with multiple building units, we leverage global relations (i.e., graph edges) among all the building units to facilitate our prediction. For instance, one does not want to have houses to fully enclose a golf court. In this way, our graph generation network learns implicit global constraints from the existing graph and understands how to add new graph edges following such ineffable rules. To account for the exclusive units/areas, we extract the essential visual clues of the input scene from the top view of the site image through a convolutional net and fuse them into the graph generation network. 
Concretely, we construct two data structures as the network inputs. One is the top-down rendered scene image (with the detailed exclusive units labeled area description), and the other is the scene graph. Our graph generation network takes as input the scene graphs and integrates the corresponding visual clues learned from the scene image to learn the global relations of nodes, and mimic constructing new edges. The scene graph does not contain the building's visual geometry semantics, nor can it describe the forbidden area in the scene. To this end, we introduce a global graph-image feature matching loss to enable the awareness of the scene geometry during graph generation. The proposed visual context-aware global relation learning network can precisely describe the geometric and topological semantics of the input scene. The auto-regressive generative mode within the network can effectively model the edge distribution from the existing nodes to the future nodes. Finally, we infer the recommended location for guiding the placement from the learned edge distribution.


We have qualitatively and quantitatively evaluated our method on a residential housing dataset collected from a commercial game. The results show that our  method can effectively model the global spatial rules in the layout of building components. With the extracted visual clues, our network effectively avoids suggestions in the forbidden areas and collisions with existing buildings. The perceptual study and the quantitative evaluation results demonstrate that our generated location maps yield meaningful and instructive guidance for the players to place new building units.


\section{Related Works}
\label{sec:relatedwork}
    \paragraph{Residential Scene Layout Synthesis.} 
    Residential scene layout synthesis plays an important role in various domains, such as game designs and architectural layouts. With the emergence of large scene datasets, more deep learning based models are proposed to address the layout generation problem. DeepSynth~\cite{wang2018deep} and FastSynth~\cite{ritchie2019fast} introduce iterative generation methods to synthesize new indoor layouts with representing the unstructured input as top view rendered images. In GRAINS~\cite{li2019grains} the input is represented as a tree structure and a recursive auto-encoder network is introduced to learn and sample the layout hierarchies.~\cite{zhang2020deep} represents the input as both arrangement matrix and rendered images and generates scenes in an attribute-matrix form with a generative adversarial network. PlanIT~\cite{wang2019planit} proposes a two-stage method with first generating a layout plan encoded as a relation graph and then instantiating the plan through an autoregressive convolutional generator based on the rendered images. In~\cite{zhang2019stylistic}, a stylistic GAN is proposed to model the relationship between the style distribution and the enhancements for 3D indoor scenes. A novel evaluation method is also introduced by \cite{liu2019qualitative} to evaluate the synthesized 3D indoor scenes qualitatively. In addition to the work mentioned above on the indoor layout generation, a few researchers have also proposed some techniques working on floor layout design. For example,~\cite{wu2019data} proposes a two-stage method to iteratively locate rooms and walls given an input boundary while \cite{hu2020graph2plan} introduces an interactive solution in which users can specify some constraints during planning. In~\cite{nauata2020house}, they propose a convolutional message passing network named House-GAN that takes as input a bubble diagram and outputs the house layout with axis-aligned bounding boxes. Unlike these tasks of indoor layout and floor plan design, our work focuses on outdoor home planning, and specifically, we aim to
    suggest locations for the new buildings.

\paragraph{Graph Generation Networks.} Graphs are natural representations of information in many areas, such as biology, engineering, and social sciences. Traditional techniques, such as~\cite{bollobas2001random, leskovec2010kronecker, margaritis2003learning, leskovec2007graph},  are based on hand-engineered graph priors that adhere to a pre-decided distribution, thus the learned generative models do not have enough capacity to represent the graph structures contained in the observed data. Inspired from recent advances in deep generative models in computer vision~\cite{wang2019generative, kingma2019introduction, kobyzev2019normalizing} and natural language processing~\cite{radford2019language,  brown2020language}, recent techniques have shifted towards a learning-based approach and have made significant progress.~\cite{simonovsky2018graphvae} proposes a VAE based graph generation model to learn to translate a latent continuous vector to a graph that can generate a graph matrix at once. However, different node ordering would lead to different graph matrices for the same graph structure, making the learning process difficult. In~\cite{li2018learning}, a message passing method is introduced to express probabilistic dependencies between nodes and edges within a graph, but the message is passing on every single edge, leading to a complex training process. GraphRNN~\cite{you2018graphrnn} proposes a hierarchical RNN framework to generate nodes and edges alternately. It also proposes a BFS node ordering scheme to improve scalability. To speed up the generative velocity, GRAN~\cite{liu2019graph} employs an efficient framework to generate one block of edge connections between nodes at a time. Inspired by GRAN~\cite{liu2019graph}, we propose a graph generation stream in our framework to learn the edge distribution between existing scene building units and the new units.

\section{The Dataset}
\label{sec:dataset}
We collected near 150K residential garden plans designed by players from a popular online game, which provides a large area of $165~grid \times 183~grid$ ($1~grid = 64 pixels$) and multiple building units of different sizes. Many players are novices to home design and landscaping, or they simply do not want to spend time on it. Some home designs are more like a collection of random building units. Significant efforts have been devoted to clean up the dataset. We first rendered all the designs into images and randomly picked 30K out of them. Those images were sent to an annotation team, consisting of trained professionals. Each image will be labeled with five grades, and a ResNet50 was trained with those manually annotated labels. Another about 30K designs that fell into the top three grades were automatically picked out by the trained model. Afterwards, the annotation team re-assessed machined-graded designed, and we kept ones labled in the top three grades.  
After this processing, our dataset contains about 28K designs, with $276$ building units per sample on average. There are $381$ different building units in total, including $280$ infrastructure units (e.g., walls, doors etc.), $101$ architectural units, and one forbidden unit that can be any shape (i.e., the pool). 

\subsection{Relation Graph Extraction}
\label{sec:graph}
We convert the scene into a directed relation graph $\mathcal{G=(V, E)}$. In this graph, nodes $\mathcal{V}$ denote scene units, which also have a spatial coordinate. Edges $\mathcal{E \subseteq V \times V}$ represent the spatial relations between nodes. 

\paragraph{Edges.} 
\label{sec:edges}
In order to encode the arrangement relations between the components, the spatial relationship is described with four direction types i.e., \textit{front}, \textit{back}, \textit{right}, \textit{left} and four distance types, \textit{next\_to}, \textit{adjacent}, \textit{proximal}, \textit{distant}, resulting 16 spatial edge types in total. To model the geometric relationship between units in more details, we also detect six edge alignment attributes, namely \textit{left~side}, \textit{vertical~center}, \textit{right~side}, \textit{top~side}, \textit{horizontal~center}, and \textit{bottom~side}. 
To extract spatial edges for node $v_i$, we first raycast from the four sides of its oriented bounding box on the $xy$ plane, and then detect intersection with other nodes. For an intersecting node, an edge is added to the graph from $v_i$ to that node if the node is visible from $v_i$ with more than $15\%$ on one side, and the directions are defined in the coordinate frame of node $v_i$. We set the distance label based on the distance between the two nodes' oriented bounding boxes: \textit{next\_to} if $distance = 0$, \textit{adjacent} if $0 < distance \leq 30$, \textit{proximal} if $30 < distance \leq 80$ and \textit{distant} otherwise. The alignment attributes are added if there is an edge connecting two nodes. 
For clarity, we only show one edge between two nodes. In fact, once one edge is detected between two nodes, we will add another edge between them (opposite direction, same distance).


\paragraph{Nodes.} An obvious strategy is to to represent a building unit as a node in the relation graph. However, as one layout design in our dataset contains about 276 different units (most of them are infrastructural units), doing so leads to an over-complicated graph. To this end, we simplify the relation graph by merging multiple infrastructure units to one node. Two units can be merged if they satisfy all the following conditions: 1) they are in the same category and have the same orientation; 2) they have the same height and are aligned in the $x$-axis or the same width and aligned the $y$-axis; 3) they are next to each other and are completely visible to each other. After merging, the number of nodes in the graph is reduced to an average of 63 with the primary information preserved.

\paragraph{Attributes.} We assign attribute vectors to each graph node and edge to encode the geometrical/semantic information of the corresponding scene. Specifically, for a node ${v_i \in \mathcal{V}}$, its attribute vector is defined as ${\widetilde{v_i} = [l_i^T, o_i^T]}$, where $l_i \in \mathbb{R}^{|D|}$ is the one hot encoded vector of the label, and $|D|$ is the number of the unit labels.  ${o_i} \in \mathbb{R}^{4}$ is the oriented bounding box of the unit on the $xy$ plane. For an edge ${e_k \in \mathcal{E}}$, its attribute vector is defined as ${\widetilde{e_k} = [t_k^T, d_k^T, m_k^T]}$, in which $t_k\in \mathbb{R}^{16}$ is the one hot encoded vector of the edge type (16 edge types in total). $t_k\in \mathbb{R}^{1}$ is the distance between two nodes. $m_k\in \mathbb{R}^{6}$ is the alignment vector of the edge.

\subsection{Top-down View Representation}
\label{sec:top-down}
We convert the 3D residential home design into a 2D layout with a top-down orthographic depth render. Doing so brings several benefits. First, since the forbidden area in the design can be in any shapes, it is difficult to represent it as a node in the graph. Instead, rendering it into a spatial image can provide detailed shape information to the network. Second, although the design is in 3D, most building units are arranged in 2D. The top-view rendering better reveals spatial outline of the design. Following~\cite{wang2018deep}, this rendering maps a $165~grid \times 183~grid$ area to a $512\times512$ image.

\section{Our Method}
\label{sec:method}
We propose a visual context-aware graph generation model to learn the edge distribution for the possible building. Our model consists of two streams: one is a ConvNet that learns detailed semantic information of each unit from the rendered image; the other stream is a graph generation network that takes as input the relation graph and fuses the visual clues learned from the ConvNet and outputs the edge distribution based on the existing graph for the possible building unit. 
\begin{figure*}[t!]
  \centering
  \includegraphics[width=\linewidth]{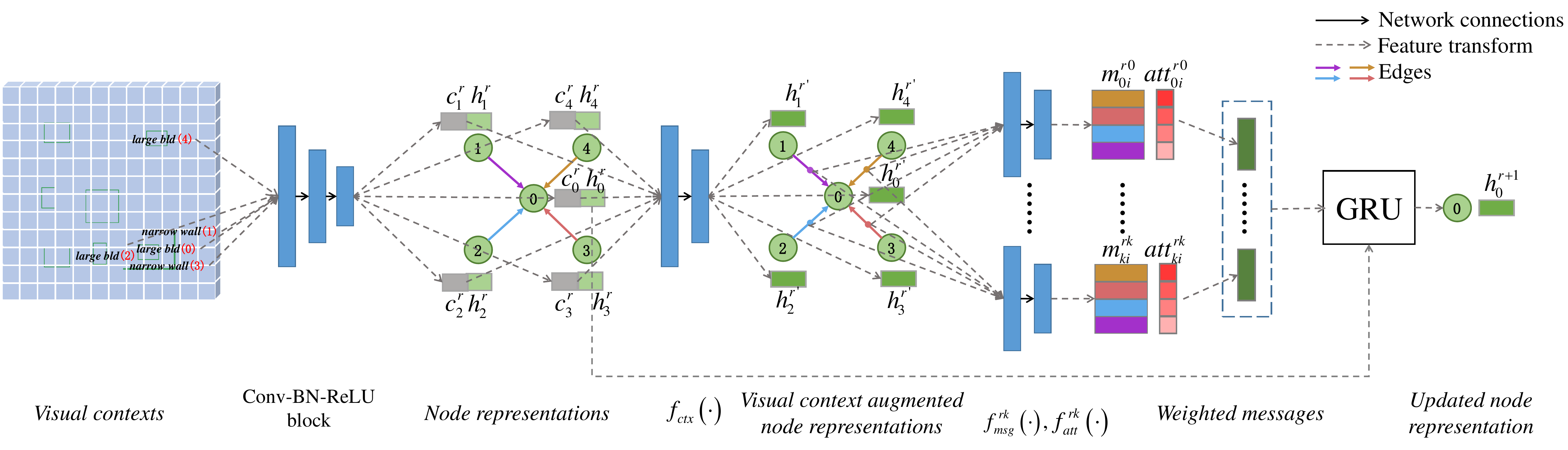}
  \caption{An overview of our proposed visual context-aware attentive message passing for the $r$-th round. This is a toy example with five building units and four edges to illustrate a single message passing iteration (``large bld'': ``large building'').}
  \label{fig:message-passing}
\end{figure*}

\subsection{Visual Context Extraction} 
We extract transformed visual features from the rendered images with a ConvNet. It is known that low-level features from a ConvNet characterize the details of local regions, and high-level features represent the global structural information of the input image. In our framework, we produce the transformed visual features using FPN based object detector~\cite{lin2017feature} in a multi-stage manner. We crop the visual feature of each building unit from the feature pyramids $\{C_1, C_2, C_3, C_4\}$ through the ROIAlignLayer~\cite{he2017mask}. Each cropped visual feature is then transformed into a fixed dimensional visual clue through a convolutional block and finally integrated into the corresponding node features in the graph relation learning network to make the learning process visual context-aware (Figure~\ref{fig:message-passing}). The convolutional block is a Conv-BN-ReLU block with a kernel size $3\times3$.

\subsection{Context-Aware Global Relation Learning}
With the extracted visual clues and the relation graph as input, our global relation learning model outputs the edge distribution between the existing nodes and the possible node. Inspired by~\cite{liao2019efficient}, we encode the edges of the relation graph $\mathcal{G=(V, E)}$ with a label weighted adjacency matrix $A$. For each edge ${(i, j) \in \mathcal{E}}$, $A_{ij} = t_{ij}$, where $t_{ij} \in T$ is the edge label, and $T$ is the edge type set. Each row vector $a_i \in A$ is interpreted as a connectivity feature of node $v_i$ representing connected relations between $v_i$ and other nodes in the graph. We learn an edge distribution $P(a_{|V|+1}|\mathcal{G})$, which samples connectivity features of relations between the new node and existing nodes in the graph. In our experiment, we only model the edge distribution from the previous nodes to the new node, which can easily infer the opposite relations. In the following, we describe how to learn edge distribution in detail. More implementation details of the network structure are provided in the supplementary material.


\paragraph{Graph Node Initialization.} 
We first translate the adjacency matrix $A$ into an one-hot matrix $\widetilde{A} $ of size $|V| \times (|T|+1) \times |V|$ with $\widetilde{A}[i, A_{ij}, j] = 1$. All the node features $\widetilde{a_i} \in \widetilde{A}$ are padded with zeros to the max dimension of the adjacency matrix in the whole dataset (443 in our dataset). Together with the node attribute vector $v_i$, the node representation is initialized as:
\begin{equation}
\label{eqn:01}
h^0_{i}=f_{init}(\widetilde{a_i}, \widetilde{v_i}; W_{init}),
\end{equation}
where $f_{init}$ is a stacked 1D convolutional block transforming the raw connectivity features $\widetilde{a_i}$ into the latent embeddings. Then the following one-layer MLP takes as input the embeddings and node attribute vector $v_i$ and outputs a node representation $h^0_{i} \in \mathbb{R}^{I}$, where $I$ is the dimension of the node representation. For the new node representation, we set $h^0_{|V|+1}=0$ and $h^0_{|V|+1} \in \mathbb{R}^{I}$. 

\paragraph{Edge Masked Attentive Message Propagation.} 
With node features (including the node representation and the corresponding visual clues) and associated attribute vectors, stacked edge masked attentive message propagation blocks propagate the messages and update the node representations' state. For the new node, we assume that it is connected to all existing nodes with an unknown label. At the $r$-th step, we first compute the visual semantic augmented node representations for all graph nodes:
\begin{equation}
\label{eqn:02}
h^{r'}_{i}=f_{ctx}(h^{r}_{i}, c^{r}_{i}; W_{ctx}),
\end{equation}
where $h^{r}_{i}$ is the node representation, and the $c^{r}_{i}$ is the corresponding cropped visual clue. $f_{ctx}$ is a two-layer MLP with learnable parameters $W_{ctx}$ to make the output node representations aware of the corresponding visual clues. 

To propagate messages and update node representations, the multi-head attention mechanism~\cite{velivckovic2018graph} is used to weight different messages for different nodes:

\begin{align}
\label{eqn:03}
m^{rk}_{ij}&=f^{rk}_{msg}(h^{r'}_{i}, h^{r'}_{j}, \widetilde{e_k}; W^{rk}_{msg}), \\
\label{eqn:04}
ma^{rk}_{ij}&=f^{rk}_{att}(h^{r'}_{i}, h^{r'}_{j}, \widetilde{e_k}; W^{rk}_{att}), \\
\label{eqn:05}
\text{att}^{rk}_{ij}&=\frac{\text{exp}(ma^{rk}_{ij})}{\sum_{l\in\mathcal{N}(i)}(ma^{rk}_{il})}, \\
\label{eqn:06}
h^{r+1}_{i}&=f^{r}_{\text{GRU}}(h^{r}_i, \|_{k=1}^{k=K}{\textstyle\sum}_{j \in \mathcal{N}(i)}\text{att}^{rk}_{ij}m^{rk}_{ij}; w^{rk}_{\text{GRU}}).
\end{align}
Here, $K$ indicates that we use $K$ different attention mechanisms to transform the messages flowing on the edges. In the $k$-th attention mechanism, we first compute the message $ma^{rk}_{ij}$ for all triplets $[v_i, e_k, v_j]$ (where $v_i$ and $v_j$ are the two nodes of the edge $e_k$) according to Eq.~\eqref{eqn:03}. An edge masked self-attention weights on messages is then obtained (according to Eq.~\eqref{eqn:04} and Eq.~\eqref{eqn:05}) to compute a linear combination of the messages for each node. Finally, the graph node representations are updated with the concatenation of the $K$ different message combinations from $K$ different attention mechanism (according to Eq.~\eqref{eqn:06}). In our experiments, $f^{rk}_{msg}$ is a two-layer MLP with learnable parameters $W^{rk}_{msg}$, 
$f^{rk}_{att}$ is implemented as a single-layer forward neural network followed with a ReLU nonlinearity. $\mathcal{N}(i)$ indicates the neighboring nodes for each node $i$, and $w^r_{\text{GRU}}$ are the learnable parameters for GRU. We show an example of this process in Figure~\ref{fig:message-passing}.

\paragraph{Edge distribution Modelling.}
After $R$ steps of message propagation, we obtain the final node representations $h^{R}_{i}$ for each node $i$ and compute the raw messages from the existing graph to the new node $m^R_{i, |V|+1} = [h^{R}_{i}, h^{R}_{|V|+1}]$. We model the edge distribution from existing nodes to the new node $P(a_{|V|+1}|\mathcal{G})$ with a mixture of categorical model based on the raw messages: 
\begin{align}
\label{eqn:07}
&P(a_{|V|+1}|\mathcal{G})=\sum^{S}_{k=1}\alpha_{s}\prod_{1\leq j \leq |V|}\theta_{s, j, |V|+1}, \\
&\alpha = \text{Softmax}(\sum_{1\leq j \leq |V|}f_{\alpha}(m^R_{j, |V|+1}; W_{\alpha})), \\
&\theta = \text{Sigmoid}(f_{\theta}(m^R_{j, |V|+1}; W_{\theta})),
\end{align}
where $S$ is the number of the mixtures in our experiments. $\alpha$ is the mixed coefficient of $S$ dimension. $\theta$ is the learned edge probabilities of different mixtures. Both $f_{\alpha}$ and $f_{\theta}$ are implemented as a two-layer MLP, and $W_{\alpha}$ and $W_{\theta}$ are the learnable parameters. The mixture of categorical distribution provides an efficient way to capture dependence in the output distribution due to the latent mixture components.

\paragraph{Losses.} 
To learn the edge distribution from existing nodes to the new node, we define the objective function as the negative log posterior probability of the mixture model:
\begin{equation}
\label{eqn:10}
\mathcal{L}_{o} = -\sum_{z=1}^{Z}\text{log}P(a_{z, |V|+1}|\mathcal{G}_{z}),
\end{equation}
where $Z$ is the batch size. To encourage the graph generation network to perceive the global visual semantics, we add global graph-image matching loss to minimize the matching score for the graph image pair. We first obtain the two global features ($\widetilde{v_G^r}$ and $\widetilde{v_I^r}$) by averaging the corresponding node features for simplify. The matching score is defined as a cosine similarity: 
\begin{equation}
\label{eqn:11}
R(G_z^r, I_z^r) = \frac{\widetilde{v_G^r}^T\widetilde{v_I^r}}{\|\widetilde{v_G^r}\|\cdot \|\widetilde{v_I^r}\|}.
\end{equation}
Similar to~\cite{xu2018attngan}, for a batch of graph-images $\{(G_z, I_z)\}_{z=1}^Z$, the posterior probability of image $I_z$ being matching with graph $G_z$ is computed as:
\begin{equation}
\label{eqn:12}
P(G_z^r|I_z^r) = \frac{\text{exp}(\gamma R(G_z^r, I_z^r))}{\sum_{b=1}^{Z}\text{exp}(\gamma R(G_z^r, I_b^r))},
\end{equation}
and the paired symmetric loss is defined as the negative posterior probability: 
\begin{equation}
\label{eqn:13}
\mathcal{L}_{m}^r = -\sum_{z=1}^{Z}\text{log}P(G_z^r|I_z^r) - \sum_{z=1}^{Z}\text{log}P(I_z^r|G_z^r).
\end{equation}
Finally, the objective function of our model is:
\begin{equation}
\label{eqn:14}
\mathcal{L} = \mathcal{L}_{o} + \sum_{r=1}^{R}\mathcal{L}_{m}^{r}.
\end{equation}

\section{Implementation Details}
In our implementation, we first extract relation graphs and rendered images from the unstructured sites. For the ConvNet to extract the visual clues for each component, we implement it based on Detectron2 and choose ResNet50 as the backbone. The aspect ratio is set as $[0.25, 0.5, 1.0, 2.0, 4.0]$. This detection model is pretrained on our rendered scene images, and we will not change the parameters in the following training phases. The cropped features are then transformed into visual clues with the size of $1,024$ based on the convolution block. For the graph generation network, we first learn the initial node representation with a size of $512$. Then we update the node representations $4$ rounds together with the corresponding visual clues based on the stack of edge masked attentive message propagation blocks. For each message passing block, we first obtain the messages with the dimension of $128$ and then concatenate the 4-head attention mechanism output to update the node representation. We add global graph-image matching loss at every round of message passing. To model the latent dependencies between edges, we set the number of mixtures to $10$ in the edge distribution model. During the training phase, we set the batch size as $32$ and choose the Adam solver for optimization, with the initial learning rate of $lr=10^{-4}$. The model is trained on 4 TitanX 2080 GPUs. 

\begin{figure}[t!]
  \centering
  \includegraphics[width=\linewidth]{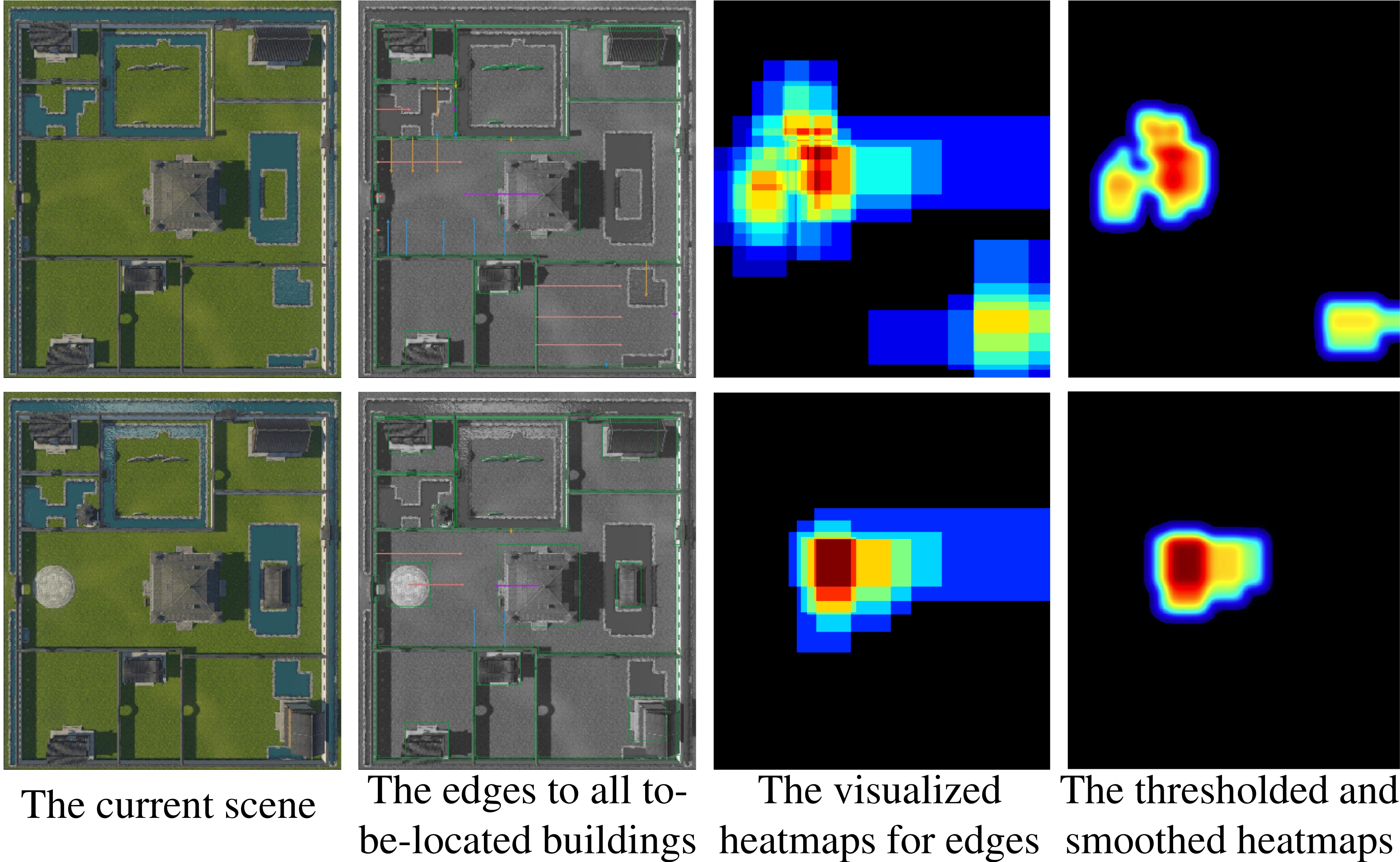}
  \caption{The visual examples of the discrete labels for evaluation.}
  \label{fig:edge-map}
\end{figure}

\section{Experiments}
We have systematically tested the proposed method. We choose 22.4K designs as the training set and the rest 5.6K are used for testing.

\paragraph{Visualization.} 
In order to intuitively evaluate our experimental results, we first convert the discrete edges from the existing scene graph to the target node for the testing set and predictions based on introduced above. Since our purpose is to recommend a location for the possible building unit, we set the default target unit size as $[24, 24]$ when visualizing edges for both the ground truth testing dataset and the predictions. During the visualization process, for the edge set $\{e\}_t$ from the current scene to the component $t$, the probability of the location the current edge points to is set as $1/|\{e\}_t|$. The final heatmap is the sum of all the locations' probability values implied by the corresponding edges. The heatmap is normalized to a maximum value of 1. In our experiment for perceptual study, we only keep the areas with the probability value greater than $0.5$ in the heatmap and smooth them with a Gaussian kernel  ($\text{kernel\_size}=5$), which better indicates our recommendation. Several examples are shown in Figure~\ref{fig:edge-map}. 

\begin{figure*}[t!]
  \centering
  \includegraphics[width=\linewidth]{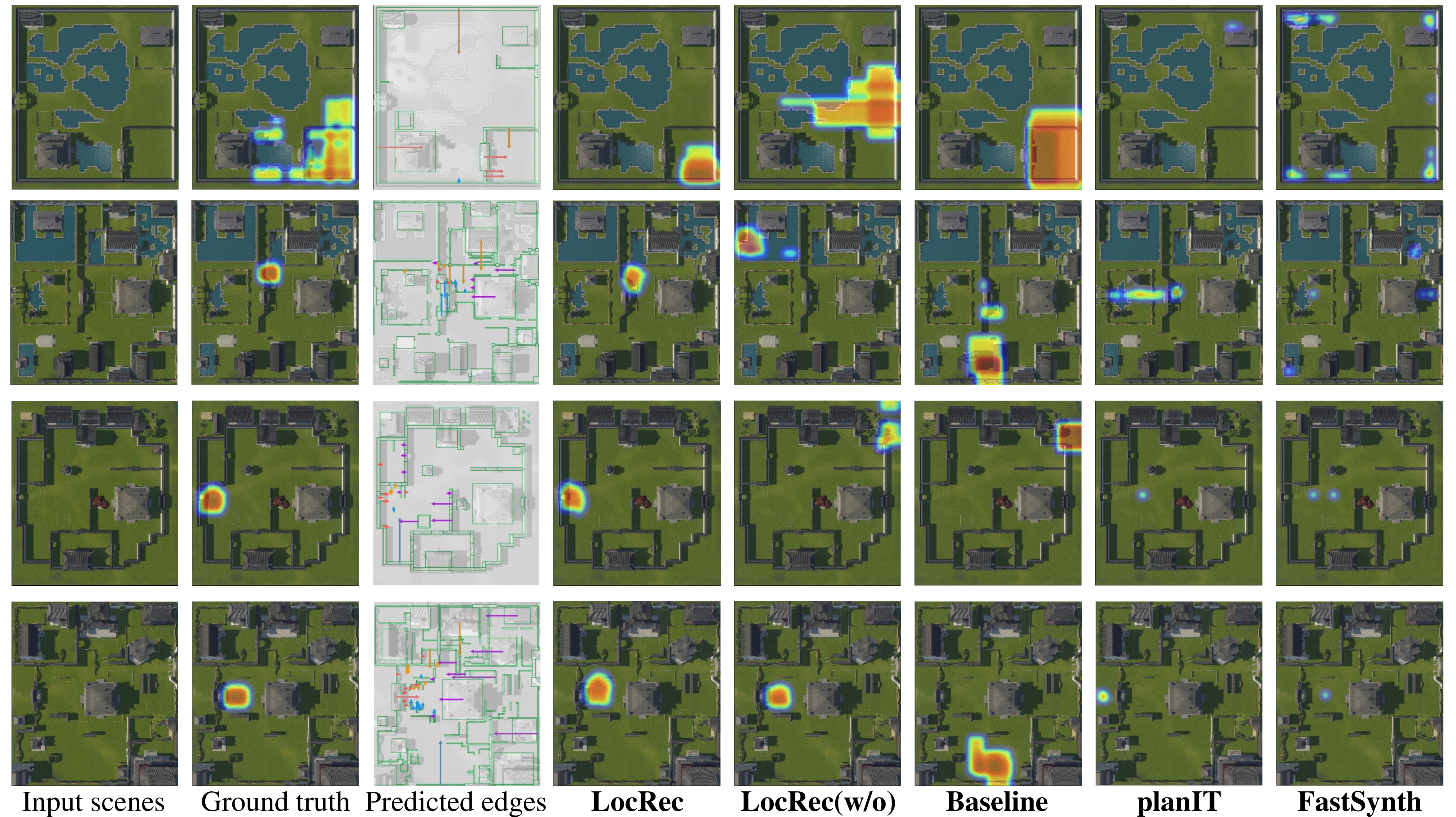}
  \caption{Location predictions using different algorithms on our testing dataset.}
  \label{fig:compare}
\end{figure*}

\paragraph{Comparisons.}
We compare our method with planIT~\cite{wang2019planit} and FastSynth~\cite{ritchie2019fast}. While those two methods are originally designed for indoor scenes, they are quite relevant to our method. For a fair comparison, we implement the \textit{partial graph completion} of planIT and add only one unit for each scene during the test. For~\cite{ritchie2019fast}, we implemented the \textit{Object Location} and choose the predicted location map indicated by the ground truth label. 
The heatmaps generated in our method are visualized from discrete edges. It is more coarse-grained than a pixel-level prediction. Therefore, we enlarge the areas of position with a probability value higher than the mean probability value for a fair comparison. We enlarge the area for each position with a size of $[24, 24]$ and centroid of itself. Then the heatmap is thresholded and smoothed as the above visualization method. We also compared our method with two degraded variants. The first variant is \textbf{Baseline}: we implemented the GRAN model~\cite{liao2019efficient} with 5 GNN layers and tested on our dataset. In this model, the relation graph is the only input. The second variant is \textbf{LocRec(w/o)}: we implemented with the proposed model without the global graph-image matching loss during the training process. We denote our model as \textbf{LocRec} in the benchmark reports.

\paragraph{Quality Metrics.}
We provide two types of metrics to evaluate the quality of predicted location maps. First, we define two criteria to evaluate the visualized heatmaps. For the ground truth heatmap $ht_r$ and the corresponding predicted heatmap $ht_p$,  we first calculated the mask $m$ of the intersection of the two non-zero areas, then the $\text{f1\_score}$ on both areas and probabilities are calculated to evaluate the results. The recall and the precision score of the area are defined as $\text{ar} =\sum m / \sum(ht_r >0)$ and $\text{ap} =\sum m / \sum(ht_p >0)$, the final $\text{f1\_score}_{area}=2/(\text{ap}^{-1} + \text{ar}^{-1})$. To calculate the $\text{f1\_score}$ on probabilities, the recall and the precision score are defined as $\text{pr} =\sum \min(ht_r[m], ht_t[m]) / \sum ht_r$ and $\text{pp} =\sum \min(ht_r[m], ht_t[m]) / \sum ht_p$. The high $\text{f1\_score}$ on area indicates that our model can effectively recommend the location for the new building units, while the high $\text{f1\_score}$ on probabilities indicate that our recommend location is compact and has clear guiding significance.

Since we aim at prompting players where to place new units, we also provide a ranking choice perceptual study to rank the results generated by different methods. We provided 60 questions and invited 45 participants to rank the different results in each question. We provide two questions for each participant, ``does this heatmap clearly specify a location?'' and ``are you willing to place a building at this location?''. Participants are asked to rank the results based on their answers. We define five levels to quantify the results, and participants are not allowed to give the same rank to different results in the same question. Rank5 represents the best, while Rank1 represents the worst. A detailed questionnaire is provided in the supplementary material.


\paragraph{Experimental Results.}
Figure~\ref{fig:small-teaser} shows a location map predicted by our method, where the heatmap is directly visualized from the predicted edges. We observe that the results generated by visualizing edges occupies a relatively large area, but the central location with high probability is still obvious, which has good guiding significance. We report more results in Figure~\ref{fig:compare} (more results in the supplementary material), where the background is the input scene, and the corresponding predicted location map is imposed to the background as a heatmap. To more clearly point out the locations generated by different methods, we show the results after thresholding and smoothing in the figure. For the results generated by our method (i.e., LocRec), we also plot the generated edges. We observe that our model can effectively learn the relations contained in the scenes, and the edge set predicted from the learned distribution has a high consistency and seldom point to multiple areas at the same time, which ensures the stability of our results. 

\begin{table}[t]
\centering
\resizebox{\linewidth}{!}{
\begin{tabular}{l|cccccc}
    \hline
    Method & $\text{ar}$ & $\text{ap}$ & $\text{pr}$ & $\text{pp}$ & $\text{f1s}_{a}$ &
    $\text{f1s}_{p}$ \\
    \hline
    \textbf{LocRec} & 0.494 & 0.847 & 0.297 & 0.551 & 0.624 & 0.386 \\
    \textbf{LocRec(w/o)} & 0.463 & 0.834 & 0.283 & 0.561 & 0.596 & 0.376 \\
    \textbf{Baseline} & 0.424 & 0.802 & 0.292 & 0.450 & 0.555 & 0.348  \\
    \hline
    \textbf{PlanIT} & 0.428 & 0.778 & 0.138 & 0.522 & 0.550 & 0.219\\
    \textbf{FastSynth} & 0.352 & 0.779 & 0.0981 & 0.499 & 0.485 & 0.164\\
    \hline
\end{tabular}}
\caption{The quantitative score on the testing dataset for different methods. $\text{f1s}_{a}$: $\text{f1\_score}_{area}$; $\text{f1s}_{p}$: $\text{f1\_score}_{probability}$.}
\label{table2}
\end{table}


We find that the results generated by our method can indicate accurate (only one peak area exists in the heatmap and the area with large probability values is very compact) and collision avoidance locations, which is instructive and meaningful to locate the new building units for players. We also find that our predictions can reasonably avoid the forbidden areas and are harmonious with the existing scenes. The corresponding quantitative results are shown in Table~\ref{table2}. It can be seen that our generated locations can hit the ground truth results in the testing dataset in most cases (with $\text{f1\_score}_{area}=62.4\%, \text{f1\_score}_{prob}=38.6\%$), which shows that our approach can effectively model the sptial rules within the units in the scene. The results of the perception study (Table~\ref{table3}) also confirm that the players accept our recommendation results and are willing to place buildings in such locations in most cases (with $\text{\#Rank5}=64\%$). 

\begin{table}[t]
\centering
\resizebox{\linewidth}{!}{
\begin{tabular}{l|ccccc}
    \hline
    Method & Rank1 & Rank2 & Rank3 & Rank4 & Rank5\\
    \hline
    \textbf{LocRec} & 0.0063 & 0.086 & 0.0060 & 0.26 & 0.64 \\
    \textbf{LocRec(w/o)}& 0.18 & 0.17 & 0.18 & 0.24 & 0.24 \\
    \textbf{Baseline} & 0.16 & 0.16 & 0.44 & 0.15 & 0.081  \\
    \hline
    \textbf{PlanIT} & 0.11 & 0.29 & 0.25 & 0.22 & 0.13\\
    \textbf{FastSynth} & 0.55 & 0.17 & 0.12 & 0.079 & 0.078\\
    \hline
\end{tabular}}
\caption{The resulting scores of the perceptual study for different methods.}
\label{table3}
\end{table}


The visualization results of planIT are given in Figure~\ref{fig:compare}, the quantitative results and the perception study results are provided in Tables~\ref{table2}~and~\ref{table3}. Compared with FastSynth, planIT gives a more reasonable location, which verifies that the constraint of the extrinsic relation graphs is more conductive to us recommending a reasonable location. Because planIT relies on the local relationship of the current unit when recommending locations, it is more difficult to learn the global relations between units in the scene, the resulting recommendation locations are much worse than our result. From the perceptual study results, compared to planIT we observe that the players are more satisfied with the location recommended by LocRec. This is because the construction site has a large space, the locations and probability values recommended by planIT can be scattered. Our method is based on global relations and leads to consistent recommendations. The $\text{pr}$ score in Table~\ref{table2} also reflects this fact.


We also provide the visualization results of our method and its variants in Figure~\ref{fig:compare}. The corresponding benchmarks are shown in Tables~\ref{table2}~and~\ref{table3}. As one can see, our baseline model is effective. Compared to FastSynth, which only inputs visual semantics, an algorithm based on relation graph is more conducive to use to learn a compact location, even it may appear to conflict with other building units in the scene. After adding visual clues reasonably, the learned locations becomes more effective. Since LocRec(w/o) does not integrate the global 
graph-image matching loss into the learning process of edge distribution, the resulting locations cannot effectively avoid the forbidden areas (e.g., pools). With the local visual clues and the global graph-image matching loss for the learning process, our full model can effectively capture the detailed and global structure of the input scene, and resulting in the best location prediction.

\section{Conclusion}
We propose an effective location recommendation method based on a  visual context-aware graph generation network. This net learns the global relations between the building units. To integrate the visual clues to the learning process, a global graph-image matching loss in also designed to enable the awareness of the scene geometry during the graph generation. The experimental results show that our method can generate instructive and meaningful locations to place the possible units. Currently, our work focus on recommending one location for the next building unit. In practice, it is more convenient to recommend multiple choices for different units collectively, which clearly offers more options to the user during the customization. However, more building units require more flexibility and ambiguities during the learning. 
In the future, we plan to investigate possible solutions to solve this problem. Besides, quantitative measurement of uncertainty during learning is also worth exploring.


\bibliographystyle{aaai21}
\bibliography{bib}
\end{document}


\linenumbers

\title{In-game Residential Home Planning via Visual Context-aware Global Relation Learning -- Supplemental material}
\author{Paper ID 932 }
\maketitle

\section{Network structure}
\begin{figure*}[htb]
  \centering
  \includegraphics[width=\linewidth]{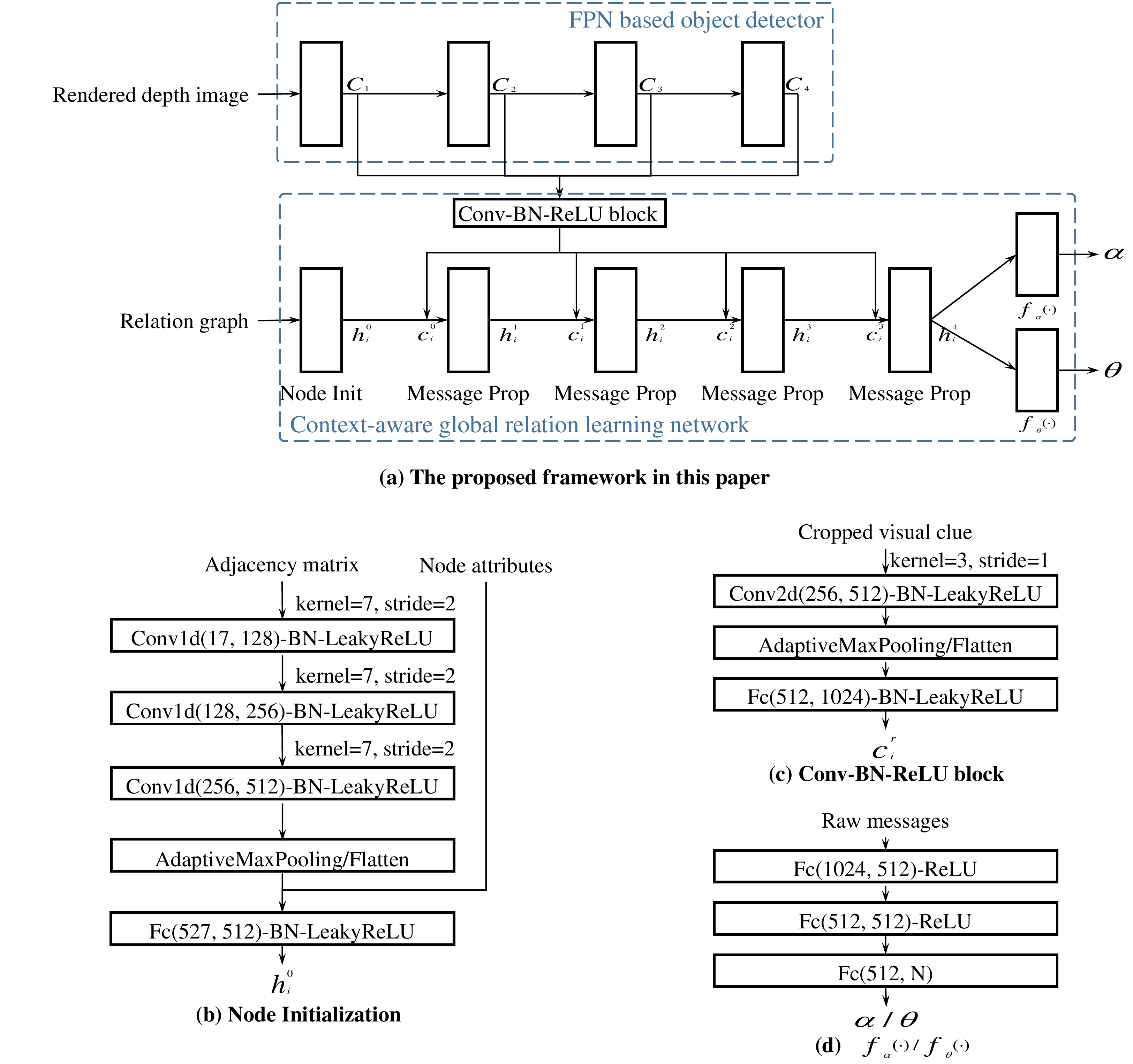}
  \caption{The implementation details of the proposed network structure.}
  \label{fig:network}
\end{figure*}

We provide detailed the network structure for our framework in Figure~\ref{fig:network}. The whole network structure is illustrated in $(a)$. As shown, the network takes as input the rendered depth image and the relation graph of the current scene, outputs parameters $\alpha$ and $\theta$ of the edge distribution. This network consists of two streams. One is the FPN based object detector to learn the visual clues of the input scene. The other stream is the graph generation network to learn the edge distribution from the current scene to a possible building unit. In the experiment, we crop the visual feature of each building unit from the feature pyramids $\{C_1, C_2, C_3, C_4\}$ through the ROIAlignLayer. Each cropped visual features is then transformed into a $1,024$-dimensional visual clue through the Conv-BN-ReLU block, which is shown in $(c)$. The context-aware relation learning stream takes as input the relation graph and the corresponding visual clues. It first transforms the relation graph into the initialized representation through the sub-network showed in $(b)$, then the node representations together with the corresponding visual clues are updated four rounds through the edge masked attentive message propagation blocks (which is depicted in the paper in detail). We provide two networks with almost the same sub-network to learn the parameters $\alpha$ and $\theta$ of the edge distribution, which is shown in $(d)$.

\section{Visualized experimental results}
\begin{figure*}[htb]
  \centering
  \includegraphics[width=\linewidth]{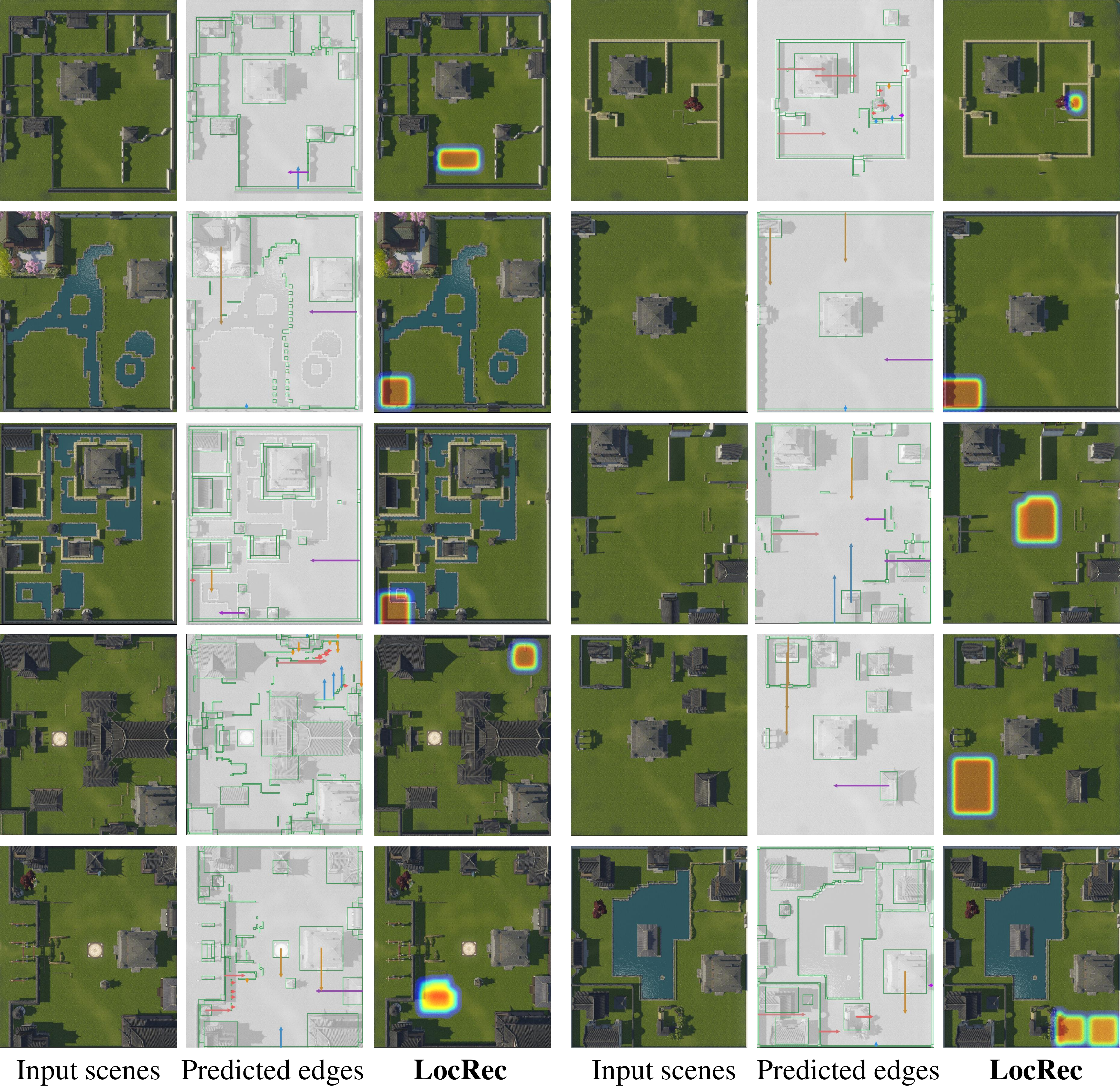}
  \caption{More visuailzed results on the test dataset with our proposed method.}
  \label{fig:compare}
\end{figure*}

More results are reported in Figure~\ref{fig:compare}, where we show the input scene, the generated edges, and the visualized result after thresholding and smoothing for each example. We observe that our model can effectively learn the scene relations, and the edge set predicted from the learned distribution has good consistency and seldom directs to multiple areas at the same time, which ensures the stability of our results. As the figure indicates, our method effectively learn the spatial rules in the placement of the building units, such as symmetrical placement (the results showed in the second row). We observe that the results generated by our method can reasonably avoid the forbidden areas and are harmonious with the existing scenes.

\section{User study}
\begin{figure*}[htb]
  \centering
  \includegraphics[width=\linewidth]{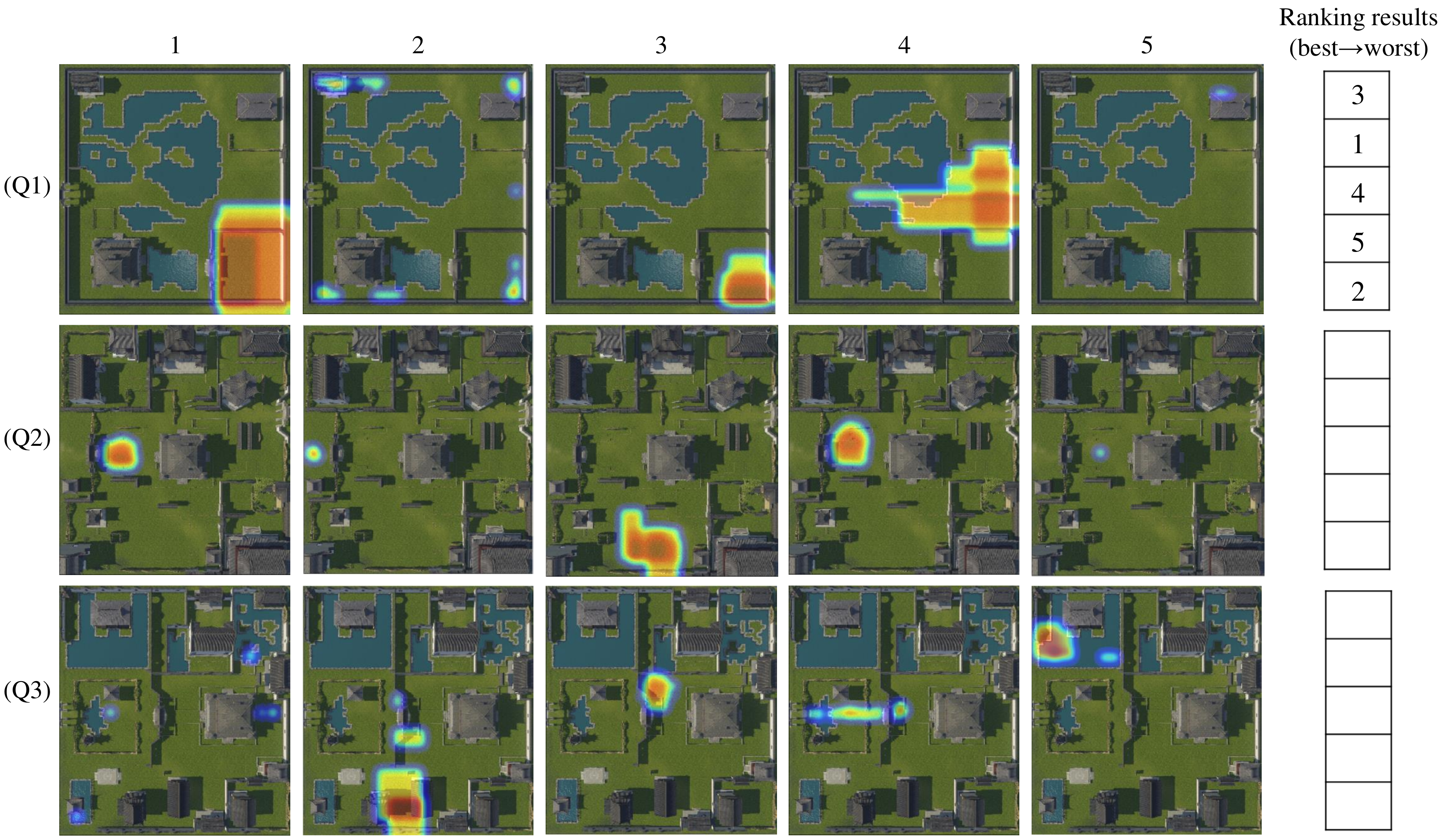}
  \caption{Three question examples in our questionnaire}
  \label{fig:examples1}
\end{figure*}

In our questionnaire, we listed 60 questions, some of which are given in Figure~\ref{fig:examples1}. The results generated by different methods are randomly showed in each question. We invited 45 participants to rank different results. For each participant, we provide two questions: ``does this heatmap clearly specify a location?'' and ``are you willing to place a building at this location?''. Participants are asked to rank the results based on their answers from best to worst. They are not allowed to give the same rank to different results in the same question. All answers need to be filled in the empty list on the right side of the question, just like $\text{(Q1)}$.

